%% file: main.tex
\documentclass[sigconf, language=french,
language=german, language=spanish, language=english,nonacm]{acmart}
\AtBeginDocument{%
  }

\begin{document}

\title{A Real-Time Human Action Recognition Model for Assisted Living}


\author{Yixuan Wang}
\affiliation{%
  \institution{National College of Ireland}
  \city{Dublin}
  \country{Ireland}
}
\email{x23162767@student.ncirl.ie}

\author{Cristina Muntean}
\affiliation{%
  \institution{National College of Ireland}
  \city{Dublin}
  \country{Ireland}
}
\email{cristina.muntean@ncirl.ie}

\author{Pramod Pathak}
\affiliation{%
  \institution{Technological University}
  \city{Dublin}
  \country{Ireland}
}
\email{pramod.pathak@tudublin.ie}

\author{Paul Stynes}
\affiliation{%
  \institution{National College of Ireland}
  \city{Dublin}
  \country{Ireland}
}
\email{paul.stynes@ncirl.ie}

\renewcommand{\shortauthors}{Wang et al.}

\begin{abstract}
  Ensuring the safety and well-being of elderly and vulnerable populations in assisted living environments is a critical concern. Computer vision presents an innovative and powerful approach to predicting health risks through video monitoring, employing human action recognition (HAR) technology. However, real-time prediction of human actions with high performance and efficiency is a challenge. This research proposes a real-time human action recognition model that combines a deep learning model and a live video prediction and alert system, in order to predict falls, staggering and chest pain for residents in assisted living. Six thousand RGB video samples from the NTU RGB+D 60 dataset were selected to create a dataset with four classes: Falling, Staggering, Chest Pain, and Normal, with the Normal class comprising 40 daily activities. Transfer learning technique was applied to train four state-of-the-art HAR models on a GPU server, namely, UniFormerV2, TimeSformer, I3D, and SlowFast. Results of the four models are presented in this paper based on class-wise and macro performance metrics, inference efficiency, model complexity and computational costs. TimeSformer is proposed for developing the real-time human action recognition model, leveraging its leading macro F1 score (95.33\%), recall (95.49\%), and precision (95.19\%) along with significantly higher inference throughput compared to the others. This research provides insights to enhance safety and health of the elderly and people with chronic illnesses in assisted living environments, fostering sustainable care, smarter communities and industry innovation.
\end{abstract}



\maketitle

\section{Introduction}
Assisted living has become increasingly popular among the elderly and individuals with chronic illnesses who seek a home-like environment with 24/7 access to medical and personal care \cite{Zimmerman2022}. These residences place more emphasis on autonomy and privacy compared to traditional hospitals and nursing homes, providing a higher quality of life \cite{Wang2023}. Their demand is rapidly expanding, particularly due to the aging population, with estimates that one in six people will be over 60 by 2030 \cite{who_ageing}. Monitoring health risks is an essential task in assisted living \cite{Tay2023}. However, traditional monitoring requires staff to perform periodic checks, which are often time-consuming, costly, prone to overlooking emergent threats, and disruptive to daily routines of residents. Additionally, we are facing a shortage of nurses worldwide \cite{who2024nursing}. 

Advanced computer vision technology in AI, human action recognition (HAR), has recently demonstrated remarkable capacity for recognising human actions through videos \cite{shuchang2022surveyhumanactionrecognition}, which could be a viable solution to these challenges. This research leverages advanced HAR technology to predict health risks in real time in assisted living, such as falls, staggering and chest pain, through continuous video monitoring. Falling or staggering may result from accidents, or worse, fatal diseases such as cardiovascular disease or cerebral infarction. Falls may lead to serious injuries, such as fractures and paralysis which can be extremely dangerous for the elderly people \cite{Clancy2015}. Moreover, chest pain is dangerous as it may indicate heart attacks or angina, which require immediate medical attention \cite{harvard_chest_pain}. Prompt alerts can significantly increase the likelihood of survival for affected individuals, by enabling timely intervention and treatments. 

Real-time prediction of these critical scenarios in assisted living can contribute to several of the UN Sustainable Development Goals (SDGs), especially the Target three \cite{sdg_goal3} that addresses the reduction of health risks and premature death from non-communicable diseases through early warning and diagnosis. It also supports building inclusive, safe and sustainable cities and communities (SDG 11) \cite{sdg_goal11} and advancing industry innovation and resilient infrastructure (SDG 9) \cite{sdg_goal9}. Harnessing this cutting-edge technology to identify critical scenarios in assisted living can potentially enhance safety and comfort in assisted living, alleviate healthcare staff shortages, support age in place, and contribute to societal sustainable development. 

The aim of this research is to investigate to what extent a human action recognition (HAR) model can predict life-threatening scenarios such as falls, staggering, and chest pain to improve assisted living environments. To address the research question, the following specific sets of research objectives were derived:
1. Investigate state-of-the-art video-based deep learning approaches to predict and classify human actions.
2. Design an architecture for a real-time human action recognition model for assisted living to predict falls, staggering, and chest pain from normal scenarios, enabling timely notifications. 
3. Implement deep learning models.
4. Evaluate the deep learning models based on model performance, inference efficiency, model complexity, and computational cost.

The major contribution of this research is a novel real-time human action recognition model that combines a deep learning HAR model with a live video prediction and alert system to predict dangerous scenarios for assisted living. In order to identify the optimal deep learning model, this research compares results of transfer learning with SlowFast, I3D, TimeSformer and UniformerV2, based on confusion matrix, class-wise and macro recall, precision and F1 score, inference throughput, number of parameters, FLOPs, and training time. The live video prediction and alert system aims to process live videos from assisted living environments, make inferences and send notifications regarding critical scenarios. 

This paper reviews AI advances in health risk prediction and human action recognition models in Section Two: Related Work. It then discusses the methodology for developing deep learning HAR models in Section Three. The architecture design of the real-time human action recognition model for assisted living is discussed in Section Four. Section Five demonstrates implementation of the deep learning models. The evaluation results of the deep learning models are presented and discussed in Section Six. Section Seven concludes the research and outlines future work.

\section{Related Work}
\subsection{AI in Health Risks Prediction}
Advanced AI technology can be leveraged to predict health risks, including accidental incidents, such as falls, choking, injuries, and burn, and non-accidental health risks, such as Alzheimer disease, dementia \cite{GarciaConstantino2020}, and behavioral anomalies \cite{Dang2020}. These technologies can be broadly grouped into two categories: one is sensor-based approach, and the other is video-based approach \cite{Tay2023}. 

The sensor-based approach uses wearable and environmental sensors. For example, \cite{Vandeweerd2020} designed HomeScenes, set up in a smart home laboratory environment, to understand residents’ behavior by analyzing sensor data, such as using door sensors to record home entry and exit, bathroom sensors to detect bathroom usage, environmental sensors to record temperature, luminance and humidity changes. \cite{Dang2020} proposed a sensor fusion approach to predict cognitive health risks, such as dementia, through wearable and ambient sensors placed in a smart kitchen, to analyze behavior patterns through daily activities, such as preparing and serving coffee and tea. Sensor-based approach offers several advantages: no body images are recorded, ensuring better privacy protection, and wearable sensors are not restricted to locations. However, specifically designed delicate sensors are often costly in both experiments and implementation and can present unstable performance when the environment changes. Moreover, wearable sensors raise concerns about comfort and inconvenience in daily life \cite{Wang2023}. These shortcomings limit their implementation in predicting critical scenarios in assisted living environments in the long term. 

The video-based approach uses images or videos as input. Although there are concerns of dependence on video quality, dataset scarcity, and privacy intrusions, it offers greater comfort without direct body contact, has more predictable costs, and is more scalable compared to the sensor-based approach \cite{Tay2023}. This approach has been widely adopted by researchers for predicting health risks \cite{Tay2023}. For example, many promising studies have been conducted to predict falls using computer vision \cite{Gaya_Morey_2024}. \cite{Lin2020} proposed a framework for real-time fall detection using human skeleton data. They combined a human pose estimation model, OpenPose, with a LSTM or GRU prediction model to predict falls and achieved an accuracy of 98\%. \cite{Paul2023}, \cite{Zheng2022}, and \cite{Iksan2021} proposed frameworks to perform abnormal or fall detection by using video data collected from centralized server-based cameras, allowing for triggering alarms in case of severe situations. Specifically, \cite{Paul2023} suggested sending messages through Twilio. \cite{Zheng2022} proposed sending the plausible video to centralized computer system for human review. \cite{Iksan2021} suggested sending video clips along with the prediction results to mobile applications.

\subsection{Deep Learning Models}
AI approaches in human action recognition can be generally categorized into conventional and deep learning methods. Conventional methods often combine a feature extractor with a machine learning model, and involve handcrafted feature selection and extensive experimentation, with feature extraction strategies varying among researchers \cite{Bourke2007}. Deep learning methods autonomously learn patterns and features from data and have recently shown their powerful aptitude for recognising human actions \cite{Cha2024}.

3D convolutional neural networks (3D CNNs) were predominant, due to CNN’s powerful feature extraction capability. By adapting 2D CNNs to process 3D data, this architecture excels at capturing spatial dependencies in videos \cite{Gaya_Morey_2024}. The two-stream inflated 3D ConvNet (I3D) \cite{Carreira2018} and the two-pathway SlowFast \cite{Feichtenhofer2019slowfastnetworksvideorecognition} model are notable representatives of 3D CNNs for human action recognition.

The I3D \cite{Carreira2018} adapts a pre-trained 2D CNN to 3D by adding a temporal dimension, thus enabling the model to learn spatial and temporal features simultaneously. In detail, it inflates filters and pooling kernels from 2D to 3D. This model leverages ImageNet architecture, and after pretraining on Kinetics, it achieved the state-of-the-art accuracy on the HMDB-51 and the UCF-101 action classification datasets. Subsequently, \cite{Feichtenhofer2019slowfastnetworksvideorecognition} proposed a two-pathway SlowFast model for video recognition. It utilizes a low frame rate to extract spatial contexts, and a lightweight high frame rate to learn rapid motion changes with less spatial nuances. The SlowFast model reported state-of-the-art accuracy on Kinetics and Charades, two video action classification benchmarks, and the AVA action detection benchmark. 

Human action recognition is intuitively considered a sequence-to-sequence problem. Therefore, many researchers have adopted recurrent neural networks (RNNs) to analyze human actions. For instance, \cite{Zheng2022}, and \cite{Lin2020} utilized RNNs for fall detection. However, unlike music or languages consisting only of temporal information, human actions comprise spatial information. Although RNNs excel at handling sequential data, they are less effective at capturing spatial features. Moreover, they sometimes encounter error accumulation issues for long-term predictions, often have high computational complexity, and are not efficient in processing periodic actions \cite{Lyu20223dhumanmotionprediction}.

Transformer-based architectures have recently become a game changer, demonstrating outstanding proficiency in understanding complex human actions and capturing long-term temporal dependency in videos \cite{Vaswani2017}. For instance, \cite{bertasius2021spacetimeattentionneedvideo} proposed the TimeSformer, one of the first models to exclusively use self-attention as building blocks for video recognition, whose design was inspired by the Vision Transformers (ViT) for image classification \cite{Dosovitskiy2021imageworth16x16words}. This model decomposes videos into frame-level patches, which are linearly embedded as input token embeddings to be fed into a transformer. Despite large number of parameters, indicating exceptional learning capacity, TimeSformer has demonstrated remarkable test efficiency. Compared to 3D CNNs, which requires long optimization procedure, this model presents fast training convergence. It can be applied on longer video clips for more than one minute. TimeSformer achieved state-of-the-art performance on Kinetics 400 and 600.

More recently, \cite{Li2022uniformerunifiedtransformerefficient} implemented the UniFormer architecture that combines CNN and transformer architectures to extract spatial and temporal features. The hybrid approach reduces computational complexity. Moreover, it applies local-window self-attention mechanisms on a small portion of video to understand long-term temporal dynamics, allowing sophisticated video processing. Subsequently, \cite{li2022uniformerv2spatiotemporallearningarming} proposed UniFormerV2, which introduces novel local and global aggregators to leverage the strengths of both ViTs and UniFormer. This model achieved state-of-the-art accuracy on eight video recognition benchmarks and was the first model to report over 90\% top-1 accuracy on the Kinetics-400. However, large model size and elevated hardware requirements of attention-based architectures may limit their usage in resource-constraint devices \cite{Karim2024}. 

In conclusion, various AI technologies have demonstrated strong potential in predicting health risks. Compared to sensor-based approach, the video-based approach is more suitable in our research due to its feasibility, convenience, scalability and more controlled costs. Many state-of-the-art deep learning human action recognition models could be leveraged to achieve this research objective, including the traditionally dominant 3D CNNs, such as I3D, which benefits from image-based pretraining, and the two-stream SlowFast network, which excels at capturing motion dynamics. Furthermore, the new powerhouse, transformer-based architectures, such as the first convolution-free, self-attention-based model, TimeSformer, and the exceptional hybrid model UniFormerV2, combining transformer architecture with CNNs, could be promising candidates for predicting life-threatening scenarios in assisted living.

\section{Research Methodology}
\begin{figure*}
    \centering
    \includegraphics[width=0.98\linewidth]{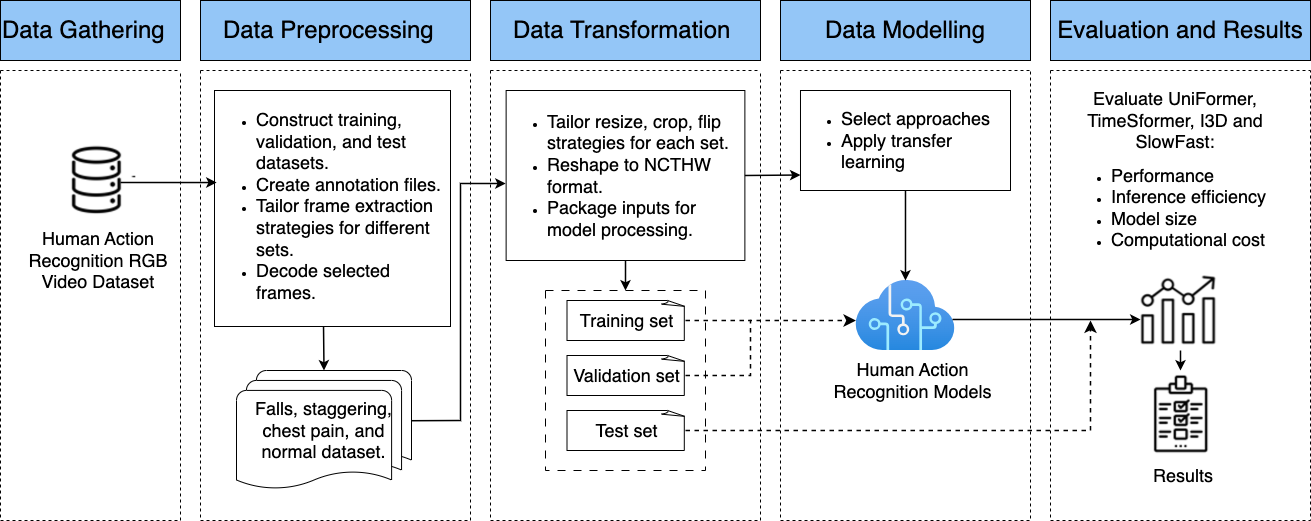}
    \caption{Research Methodology}
    \label{fig:1}
\end{figure*}

The research methodology consists of five steps namely data gathering, data pre-processing, data transformation, data modelling, evaluation and results as shown in Figure \ref{fig:1}.

The first step, Data Gathering involves identifying appropriate video data for human action recognition. This study utilised the NTU RGB+D Action Recognition Dataset \cite{shahroudy2016nturgbdlargescale}, which provides RGB video samples, depth maps, and 3D skeletal data and comprises 60 action classes, with 40 classes categorised in “Daily Actions”, 9 classes in “Medical Conditions”, and 11 classes in “Mutual Actions”. To predict life-threatening scenarios in assisted living environments, the dataset should contain all predicted classes and a “Normal Scenario” class as a contrasting group. All RGB video samples of Falling (948 videos), Staggering (948 videos), and Chest pain (948 videos) in the “Medical Conditions” category were selected to represent dangerous scenarios. The “Normal Scenario” class was formed by randomly selecting 80 videos from each of the 40 classes in the “Daily Actions” category to include a larger sample size (3,200 videos) with a wide diversity of daily activities, which simulate real-life conditions. 

\begin{table}[h]
  \caption{Dataset Statistics}
  \label{tab:1}
  \resizebox{\linewidth}{!}{%
  \begin{tabular}{lcccc}
    \toprule
    Class & Total Samples & Training (75\%) & Validation (12.5\%) & Testing (12.5\%) \\
    \midrule
    Falling & 948 & 712 & 118 & 118 \\
    Staggering & 948 & 712 & 118 & 118 \\
    Chest Pain & 948 & 712 & 118 & 118 \\
    Normal & 3200 & 2400 & 400 & 400 \\
    Total & 6044 & 4536 & 754 & 754 \\
    \bottomrule
  \end{tabular}%
  }
\end{table}

The second step, Data Preprocessing involves train, validation and test splits, feature-label mapping and raw video data preprocessing. The dataset was split into training, validation and testing sets, in proportion of 75\%, 12.5\% and 12.5\% respectively. The splitting process ensured that the proportion of the four classes remained the same in training, testing and validation sets. The dataset statistics are listed in the Table \ref{tab:1}. Then, feature-label mapping was performed. An annotation text file for each set was created, listing the relative video path with its corresponding label. The raw video data preprocessing involved frame extraction and decoding sampled frames. The frame extraction strategy varies depending on different models and is detailed in the Implementation section. All the sampled frames were then decoded into tensors for further processing.

The third step, Data Transformation includes resizing, cropping, flipping and formatting. Specifically, frames were initially resized to have a consistent height of 256 pixels, preserving the aspect ratio. They were then cropped with different methods and were resized to a consistent input shape of 224x224 pixels without maintaining aspect ratio. For the training set, frames were randomly flipped horizontally with a 50\% probability. Finally, all frames were converted into NCTHW format (batch size, channels, temporal dimension, height, width), and they were packed along with their labels and metadata for model input.

The fourth step, Data Modelling, involves identifying appropriate approach, and selecting and training deep learning models. Given the limited open-source data and computational constraints in this research, a transfer learning approach with state-of-the-art human action recognition models, including 3D CNN-based models (SlowFast and I3D), a transformer-based model (TimeSformer), and a hybrid model (UniFormerV2), was adopted. Transfer learning can leverage pre-trained model weights from proven models. Although they were trained on different datasets, their weights have successfully learned essential features for video understanding, particularly in the initial layers. Transferring these weights to a new task leads to faster convergence, enhances feature extraction, and can often deliver good performance even with a limited dataset size. SlowFast, I3D, TimeSformer, and UniFormerV2, pre-trained on the Kinetics dataset \cite{kay2017kineticshumanactionvideo}, a large-scale human action recognition video dataset, were used as the base models. The detailed experimentation is discussed in the Implementation section. 

The fifth step, Evaluation and Results evaluates the performance of the four models developed with transfer learning through class-wise and overall performance, inference efficiency, model complexity and computational cost. Confusion matrix, precision, recall and F1 score are common performance metrics for multi-class classification tasks. Considering the imbalanced dataset setup, where the normal scenarios account for higher proportion compared to each critical scenario, macro metrics which average across classes (give equal weight to each class), are more appropriate compared to micro metrices, which average across instances (assign equal weight to each instance), avoiding biases coming from the dominant class. Therefore, model performance was evaluated using confusion matrix, macro and class-wise recall (accuracy), precision, and F1 score. Inference efficiency was assessed through inference throughput. Model complexity was represented using the number of parameters. Computational cost was measured using FLOPs and training time. The optimal model was then proposed to implement the real-time human action recognition model, whose architecture is outlined in the Design Specification section. 

\section{Design Specification}
\begin{figure*}
    \centering
    \includegraphics[width=0.85\linewidth]{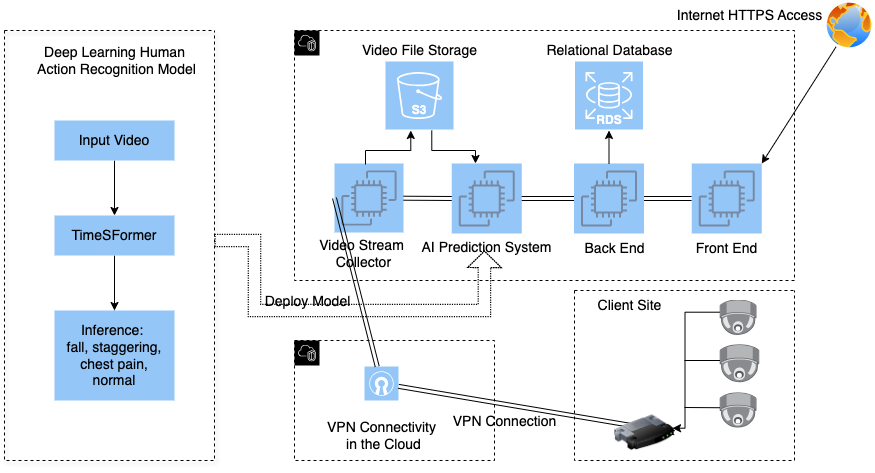}
    \caption{Architecture for the Real-time Human Action Recognition Model}
    \label{fig:2}
\end{figure*}

The real-time human action recognition model architecture combines a video-based deep learning model, and a live video prediction and alert system, as shown in Figure \ref{fig:2}. The components of the deep learning model include input video, the optimal model (TimeSformer), and inference, as discussed in section 4.1. Components of the live video prediction and alert system are discussed in section 4.2.

\subsection{Video-based Deep Learning Model}
The TimeSformer, is chosen as the deep learning model, due to its strong performance and remarkable inference efficiency among all tested models. It is recommended to be developed using GPUs, given its large model size and high computational complexity. The detailed model development setting and evaluation are presented in the paper. The optimal model selected during training stage is deployed to the AI prediction system, a component under the live video prediction and alert system, for real-time inference. 

\subsection{Live Video Prediction and Alert System }
The live video prediction and alert system consists of a video stream collector, an AI prediction system, a system backend, and a frontend component. These components should be implemented using a modular approach, which is easier for maintenance, and they communicate through API REST interfaces. 

Firstly, the video stream collector receives streaming video from live cameras in assisted living through an established VPN connection. VPN server simplifies the connection of cameras from client site to the cloud. Specifically, it is achieved by having the client site router initiate a VPN connection to the VPN server in the cloud (e.g., using an OpenVPN server). Once the connection is established, the cameras can be accessed by the video stream collector through their private IP addresses, via the router at client site. This approach minimises the setup complexity at the client site, as no extra network configuration is required. Furthermore, VPN server provides additional layer of security by encrypting video stream during transit, enhancing data privacy and protection. Secondly, the video stream collector performs a fixed-time window segmentation, which cuts the video into small chunks, each of a similar length to the videos used in dataset for model development. It will then send the video chunks to a file storage service, such as Amazon S3 cloud object storage. 

The AI prediction system contains the pretrained deep learning model for making inferences. It’s also suggested to be deployed using GPU servers, which have proven to deliver promising results in this research. The AI prediction system loads video chunks from the file storage service for preprocessing and transformation using a predefined test pipeline. The test pipeline includes steps such as frame extraction, decoding, resizing, cropping, NCTHW format conversion, and packaging inputs for model processing. Then, the AI prediction system make an inference for each segment, classifying it into one of the four scenarios: Falling, Staggering, Chest pain or Normal. 

The model continuously generates predictions and sends the inference results to the backend. The backend saves these inferences in a relational database server, such as MySQL or AWS RDS. Once the predictions contain critical scenarios, the backend software highlights the specific video chunk and sends a notification regarding the specific event. 

The information is then sent to the frontend, which hosts a basic application that the assisted living workers or family member of the residents have access to. They should review and confirm the flagged videos and take action upon it. Moreover, if a video clip is falsely predicted, it will be recorded and used for future model retraining to enhance model robustness. 

One option could be deploying the live video prediction and alert system in the cloud, given the availability of high-speed and continuous internet in assisted living environments. Cloud-based deployment often ensures none-disruptive and continuous processing, and can be assessed remotely. Cloud platforms such as AWS can be scaled up or down easily when the number of clients or video streams change. The system pays for the resources it uses, which could be cost-efficient. For instance, AWS EC2 instances can be used to host these components. Each of the component should be deployed in a separate EC2 machine. If more advanced features are preferred, AWS EKS (Elastic Kubernetes Service) can also be utilised to deploy these components. As for data storage, AWS RDS can be used for database management, and AWS S3 can be used for storing video files. 

For real-world implementation, the AI prediction system would be the most expensive component of this real-time human action recognition model, as it is deployed on GPUs, whereas other components are deployed on regular servers. The cost of using an AWS EC2 p3.2xlarge instance with an NVIDIA V100 GPU is estimated below. Assuming an inference throughput of 3.96 samples/second (TimeSformer), 10-second video chunk, and using one NVDIA V100, the total number of clients that can be monitored in assisted living is 39.6, rounded to 39 persons. An NVDIA V100 from AWS (\$3.06/h)\cite{vantage_p3_2xlarge}, incurs a cost of approximately \$2203 monthly. Therefore, at least 39 clients can be monitored 24/7 with this inference cost, which is around \$55 per client. The video chunk in our dataset is around 10 seconds; however, TimeSformer is also capable of predicting longer videos, exceeding one minute, which may worth experimenting with if such a dataset is available. Using longer video chunk reduces inference frequency and may decrease the overall inference cost with same resources. Moreover, the most recent version NVDIA A100 is more powerful than NVDIA V100, the version tested in our experiments. Using newer server might further decrease inference cost.

\section{Implementation}

The deep learning models were implemented using Python and PyTorch, with the server specifications of 64GB of RAM, an AMD RyzenTM 7 5800X processor, a Tesla V100 GPU, and two Samsung 990 PRO NVMe M.2 SSDs, each with 2TB of storage. The transfer learning was performed using the open-source deep learning ecosystem, OpenMMLab \cite{OpenMMLab}, developed by the Multimedia Laboratory (MMLab) at CUHK. OpenMMLab offers a collection of toolkits, libraries and comprehensive pre-trained model zoos for computer vision tasks. Specifically, the development version of MMAction2 \cite{MMACTION2}, providing a modular design for a video understanding framework, was utilised. 

Four models, SlowFast, I3D, TimeSformer and UniFormer, a total of six variants were evaluated in this research. TimeSformer (divided) applies spatial and temporal attention separately, whereas the joint version processes spatial and temporal attention jointly \cite{bertasius2021spacetimeattentionneedvideo}. SlowFast, with a backbone of ResNet 101, offers deeper architecture compared to the ResNet 50 backbone, with increased feature representation and computational complexity \cite{Feichtenhofer2019slowfastnetworksvideorecognition}. The detailed model information is summarised in Table 2. 

I3D and SlowFast utilise ResNet as backbone. The backbone of I3D is pretrained on ImageNet, while no pretraining is performed on the backbone of SlowFast. The backbone of TimeSformer is the Vision Transformer (ViT) \cite{Dosovitskiy2021imageworth16x16words}, pretrained on ImgeNet-1K. UniFormerV2 also utilises the ViT backbone and applies Contrastive Language-Image Pretraining (CLIP) \cite{Radford2021learningtransferablevisualmodels}. The training sampling strategy is denoted in the table as “clip length (frames sampled per clip) × temporal stride between sampled frames × number of clips per video during training”. All models are trained using 8 GPUs, with a batch size of 8 per GPU. These models are pretrained on the Kinetics dataset for varied epochs. 

During transfer learning, model's output layer was modified to four classes. A class weight of [1, 1, 1, 0.3] was set in the cross-entropy loss function for each model to address class imbalance. Training parameters were maintained as in the original model settings. The optimizer for UniFormerV2 is AdamW, all other models use SGD. Shuffling was solely applied on training data. A smaller learning rate compared to the original setup was used, combined with LinearLR and CosineAnnealingLR for dynamically adjusting the learning rate. All models were initially trained for 50 epochs, except for I3D, which was trained for 60 epochs as it continued to converge after 50 epochs. 

UniFormerV2, TimeSformer (joint), and TimeSformer (divided) required longer times for training 50 epochs, for 13.91, 16.64, and 18.15 hours, respectively, while showing faster convergence speeds. UniFormerV2 achieved a mean class accuracy of above 90\% on validation set initially at 12 epochs, using 3.51 hours. For TimeSformer, the divided variant first achieved an accuracy above 90\% in 3.23 hours at 9 epochs, while the joint version took 5.01 hours at 14 epochs.
In contrast, I3D required the least total training time, at 4.83 hours for 50 epochs and 5.80 hours for 60 epochs. It first observed a mean class accuracy above 90\% in 1.53 hours at 16 epochs. However, during multiple experiments, both SlowFast models were unable to achieve a mean class accuracy above 90\%. The training times for 50 epochs for SlowFast (ResNet 50) and SlowFast (ResNet 101) were 6.19 and 13.38 hours respectively. 

Model checkpoints from the last training epoch and the epoch with the best mean class accuracy on the validation set were selected to perform model testing on the test dataset. The best testing results were recorded as the model’s performance. These results were analysed, evaluated from multiple perspectives and compared in Section 6. The optimal model was selected for implementing the real-time human action recognition model for assisted living. 

\begin{table} [h]
  \caption{Specifications of Selected Pretrained Models}
  \label{tab:2}
  \resizebox{\linewidth}{!}{%
  \begin{tabular}{lccccc}
    \toprule
    Model & Backbone & Sampling & Batch & Epochs & Dataset \\
    \midrule
    SlowFast & ResNet 50 & 4×16×1 & 8GPUs×8 & 256 & Kinetics 400 \\
    SlowFast & ResNet 101 & 8×8×1 & 8GPUs×8 & 256 & Kinetics 400 \\
    I3D & ResNet 50 & 32×2×1 & 8GPUs×8 & 100 & Kinetics 400 \\
    TimeSformer (divided) & ViT & 8×32×1 & 8GPUs×8 & 15 & Kinetics 400 \\
    TimeSformer (joint) & ViT & 8×32×1 & 8GPUs×8 & 15 & Kinetics 400 \\
    UniFormerV2 & ViT & 8×-×1* & 8GPUs×8 & 55 & Kinetics 700 \\
    \bottomrule
  \end{tabular}%
  }
  \vspace{1mm}
  \resizebox{\linewidth}{!}{%
  \textbf{Note:} * The stride is dynamically calculated as the total frames in the video divided by the clip length.
  }
\end{table}

\section{Evaluation}
The aim of this experiment is to compare UniFormerV2, TimeSformer, I3D and SlowFast deep learning HAR models with a customised RGB video dataset to predict several critical scenarios. Through the use of transfer learning techniques, four types of models, including a total of six pre-trained models, were trained to predict falls, staggering, and chest pain out of normal activities. 

\subsection{Performance Overview}

\begin{table*}
  \caption{Performance Comparison of Models}
  \label{tab:performance}
  \centering
  \renewcommand{\arraystretch}{1.3} 
  \resizebox{\linewidth}{!}{%
  \begin{tabular}{lcccccc}
    \toprule
    Model & Mean Acc. (\%) & Throughput (/s) & Params (M) & GFLOPs & Total Train Hours & Train Hours Acc. > 90\% \\
    \midrule
    UniFormer & 95.36 & 1.28 & 114 & 143 & 13.91 & 3.51 (12 epochs) \\
    TimeSformer (joint) & 93.44 & 3.65 & 85.808 & 180 & 16.64 & 5.01 (14 epochs) \\
    TimeSformer (divided) & 95.49 & 3.96 & 121 & 196 & 18.15 & 3.23 (9 epochs) \\
    I3D & 93.43 & 0.97 & 27.232 & 33.271 & 5.80\textsuperscript{*} & 1.53 (16 epochs) \\
    SlowFast (r50) & 84.95 & 0.96 & 33.567 & 27.816 & 6.19 & - \\
    SlowFast (r101) & 86.81 & 0.37 & 62 & 97.1 & 13.38 & - \\
    \bottomrule
  \end{tabular}%
  }
  \vspace{2mm}
  \resizebox{\linewidth}{!}{%
  \begin{minipage}{\linewidth}
      \centering
      Note: * The I3D model was trained for 60 epochs, while the other models were trained for 50 epochs.
  \end{minipage}
  }
\end{table*}

Table \ref{tab:performance} shows the comparison of the six models based on mean class accuracy, inference throughput, the number of parameters, GigaFLOPs, total training hours, and the training hours required to achieve an above 90\% mean class accuracy. 

Mean class accuracy represents model’s overall accuracy among all classes. Inference throughput assesses inference speed, calculated as the number of samples processed per second on the test data, which is critical for real-time applications. The number of parameters reflects model complexity, size, and learning capacity; a higher count implies greater model complexity, and higher memory and computational power requirements. FLOPs (Floating Point Operations) evaluate computational complexity, which is the number of floating-point arithmetic operations a processor performed per second. Total training hours represent how fast a model can train on given hardware, providing insights for resource requirements and hardware benchmarking. The training hours required to achieve an above 90\% mean class accuracy indicate convergence spend, training efficiency, and practical feasibility. A long total training time suggests a computationally intensive model but does not necessarily mean that the model is inefficient in learning. For instance, TimeSformer (divided) required a long total training time for 50 epochs but a short period to achieve an above 90\% accuracy, suggesting a fast convergence speed and efficient model learning. 

The divided space-time TimeSformer achieved the highest mean class accuracy and inference throughput among all models. This variant demonstrated greater accuracy, faster inference, and quicker training convergence, compared to the joint space-time variant, despite slightly larger number of parameters, FLOPs, and total training time. Both SlowFast variants showed less satisfying performance, with less than 90\% mean class accuracy and the slowest inference speeds. 

In general, the models with transformer structures, including two TimeSformer models and UniFormerV2, have more parameters, demand higher FLOPs and total training time. TimeSFomer (divided) has the largest number of parameters, up to 196 million, contributing to its outstanding performance through more comprehensive feature learning. Notably, although TimeSformer (divided) is computationally intensive, it demonstrated the fastest inference throughput and relatively fast convergence. This characteristic has been mentioned in the original paper by the authors \cite{bertasius2021spacetimeattentionneedvideo}. In contrast, I3D has the smallest parameter count (27.232 million), the shortest total training time, and the fastest convergence speed, suggesting a less complex model with lower memory and hardware requirements, making it easier to deploy.

TimeSformer (divided) and SlowFast (r101) achieved better accuracy compared to their respective variants, therefore they are selected as representatives of their model types. The performance of the four models–UniFormerV2, TimeSformer (divided), I3D, SlowFast (ResNet101)–is further analysed and compared in the following sections.

\subsection{Confusion Matrix}

\begin{figure*} [h]
    \centering
    \includegraphics[width=0.78\linewidth]{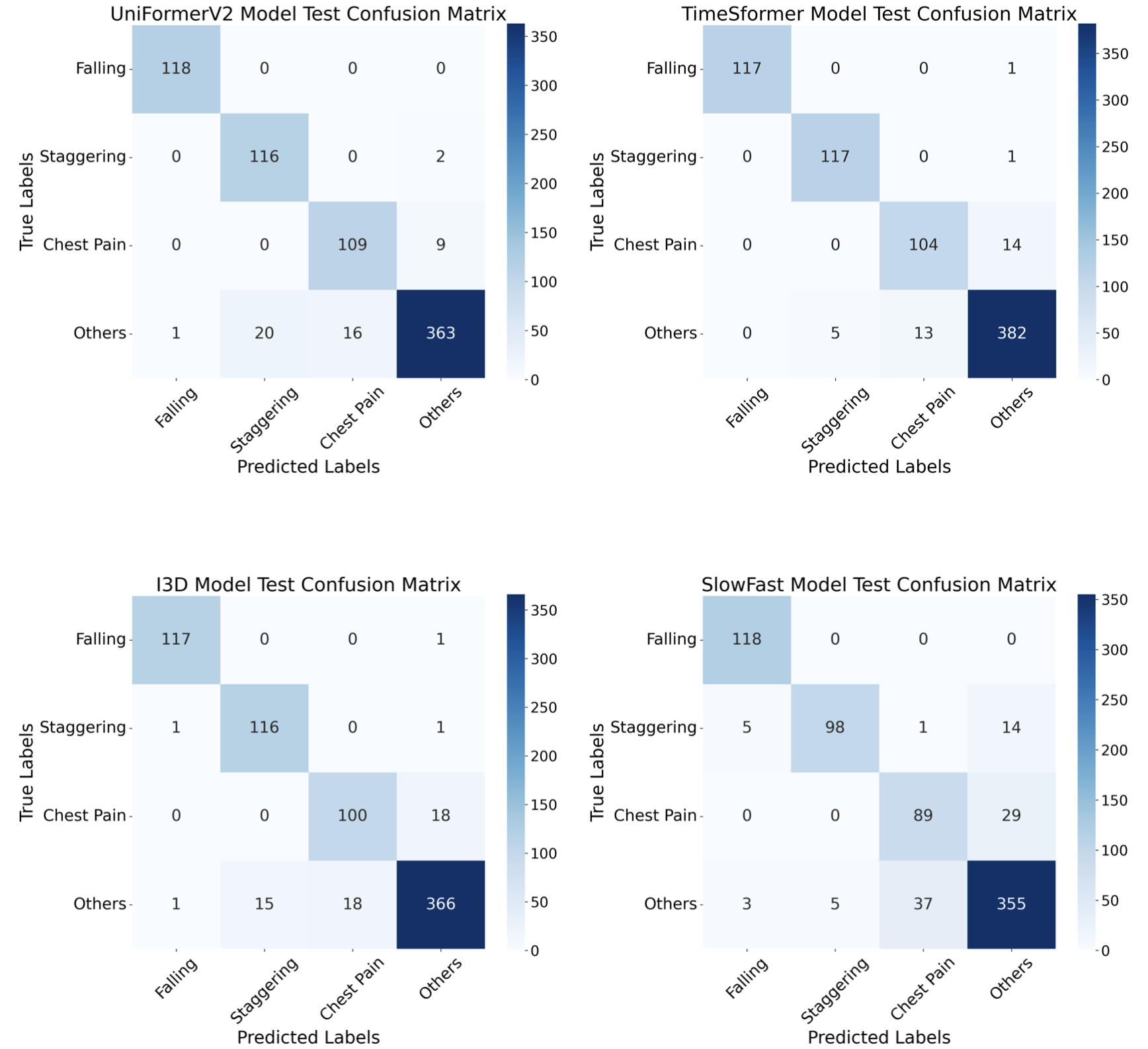}
    \caption{Confusion Matrix for the Four Models}
    \label{fig:confusion}
\end{figure*}

Figure \ref{fig:confusion} presents the confusion matrix for UniFormerV2, TimeSformer (divided), I3D, and SlowFast (ResNet101) on the test data. Confusion matrix is widely used to assess classification results, identify model weakness, and calculate other performance metrics such as recall, precision, and F1-score. In each confusion matrix, rows are “true labels” and columns are “predicted labels”. Diagonal values are the correctly predicted instances, known as true positive. The sum of all cells except for the row and column where the class is located, is true negatives, meaning correctly predicted instances of “not this class”. 

The sum of all cells in the same row, excluding the diagonal value, is the false negative for that class. This indicates a risk of missing critical scenarios, which may lead to injuries or severe consequences. UniFormerV2 and SlowFast did not encounter false negatives for the falling (0 case) class. For predicting staggering, TimeSformer recorded the fewest false negatives (1 case). For chest pain, UniFormerV2 had the fewest false negatives (9 cases). On the contrary, SlowFast incurred the highest false negatives for staggering (20 cases) and chest pain (29 cases). Overall, there are less false negatives in predicting falls, more in predicting staggering, much more frequent in predicting chest pain across all models. 

The sum of all cells in a column, excluding the diagonal value, is the false positive for that class, representing a false alarm, which can be disruptive and costly. TimeSformer observed the fewest false positives in predicting all three critical scenarios: falls (0 case), chest pain (5 cases), and staggering (13 cases). In comparison, SlowFast had the highest false positives in predicting falls (8 cases) and chest pain (38), while it faced the fewest false positives along with TimeSformer for predicting staggering (5 cases).

\subsection{Class-wise and Macro Performance Evaluation}

Model performance was evaluated using accuracy, recall, precision, and F1 score. Class-wise performance was assessed by calculating these metrics for each individual class. In multi-class classification, class-wise accuracy for a single class is calculated the same as recall, measuring the proportion of instances in one critical scenario that the model accurately predicted out of all instances in that class. Precision measures the model’s capability to predict instances of each dangerous scenario accurately. F1 score is the harmonic mean of precision and recall, measuring the overall model performance. The class-wise accuracy/recall, precision and F1 score for class “i” are calculated using formular 1, 2, and 3.

\begin{flushleft}
\begin{align}
  \text{Accuracy}_i \ / \ \text{Recall}_i &= \frac{\text{True Positive}_i}{\text{True Positive}_i + \text{False Negative}_i} \\
  \text{Precision}_i &= \frac{\text{True Positive}_i}{\text{True Positive}_i + \text{False Positive}_i} \\
  \text{F1 Score}_i &= \frac{2 \times \text{Precision}_i \times \text{Recall}_i}{\text{Precision}_i + \text{Recall}_i}
\end{align}
\end{flushleft}

High accuracy, recall, and precision are preferred. A higher recall indicates a lower risk of failing to predict a danger; a higher precision represents a lower probability of producing false alarms; and a higher F1 score reflects stronger and more balanced performance between recall and precision.

A comparison of the class-wise performance among four models is summarised in Tables \ref{tab:falling}, \ref{tab:staggering}, and \ref{tab:chestpain}. It is seen that both UniFormerV2 and SlowFast excelled in recall (100\%) for predicting falls. Overall, UniFormerV2 showed the top performance, with the highest F1 score of 99.58\% in predicting this class, and TimeSformer demonstrated the best precision (100\%). For predicting staggering, TimeSformer achieved the highest F1 score (97.50\%), recall (99.15\%), and precision (95.90\%). All four models were less effective in predicting chest pain, with UniFormerV2 showing the best F1 score of 89.71\% and recall of 92.37\%, and TimeSformer presenting the best precision (88.90\%). Class-wise analysis suggests a model's capacity of predicting certain classes. If the goal is to predict specific actions, these results can offer valuable insights.

\begin{table}[h]
  \caption{Performance Comparison for Predicting Falls}
  \label{tab:falling}
  \resizebox{\linewidth}{!}{%
  \begin{tabular}{lcccc}
    \toprule
    Metric & UniFormerV2 & TimeSformer & I3D & SlowFast \\
    \midrule
    Recall & \textbf{100\%} & 99.15\% & 99.15\% & \textbf{100\%} \\
    Precision & 99.16\% & \textbf{100\%} & 98.32\% & 93.65\% \\
    F1 Score & \textbf{99.58\%} & 99.57\% & 98.73\% & 96.72\% \\
    \bottomrule
  \end{tabular}%
  }
\end{table}

\begin{table}[h]
  \caption{Performance Comparison for Predicting Staggering}
  \label{tab:staggering}
  \resizebox{\linewidth}{!}{%
  \begin{tabular}{lcccc}
    \toprule
    Metric & UniFormerV2 & TimeSformer & I3D & SlowFast \\
    \midrule
    Recall & 98.31\% & \textbf{99.15\%} & 98.31\% & 83.05\% \\
    Precision & 85.29\% & \textbf{95.90\%} & 88.55\% & 95.15\% \\
    F1 Score & 91.34\% & \textbf{97.50\%} & 93.17\% & 88.69\% \\
    \bottomrule
  \end{tabular}%
  }
\end{table}

\begin{table}[h]
  \caption{Performance Comparison for Predicting Chest Pain}
  \label{tab:chestpain}
  \resizebox{\linewidth}{!}{%
  \begin{tabular}{lcccc}
    \toprule
    Metric & UniFormerV2 & TimeSformer & I3D & SlowFast \\
    \midrule
    Recall & \textbf{92.37\%} & 88.14\% & 84.75\% & 75.42\% \\
    Precision & 87.20\% & \textbf{88.89\%} & 84.75\% & 70.08\% \\
    F1 Score & \textbf{89.71\%} & 88.51\% & 84.75\% & 72.65\% \\
    \bottomrule
  \end{tabular}%
  }
\end{table}

\begin{table}[h]
  \caption{Macro Metrics}
  \label{tab:macro}
  \resizebox{\linewidth}{!}{%
  \begin{tabular}{lcccc}
    \toprule
    Metric & UniFormerV2 & TimeSformer & I3D & SlowFast \\
    \midrule
    Macro Recall & 95.36\% & \textbf{95.49\%} & 93.43\% & 86.81\% \\
    Macro Precision & 92.18\% & \textbf{95.19\%} & 91.61\% & 87.02\% \\
    Macro F1 Score & 93.61\% & \textbf{95.33\%} & 92.45\% & 86.76\% \\
    \bottomrule
  \end{tabular}%
  }
\end{table}

To assess overall performance, macro metrics, which are the averages of these metrics among all classes, were adopted. In contrast, micro metrics, averaging across all instances, were not adopted, as they are biased towards the dominant class, which is not suitable for this research due to the differing sample sizes across classes. A comparison of macro metrics among all models is presented in Table \ref{tab:macro}.

Macro metrics indicate that TimeSformer (divided) is the top-performing model regarding macro F1 Score (95.33\%), recall (95.49\%), and precision (95.19\%). UniFormerV2 and I3D ranked at the second and third position for these metrics.

\begin{figure}[h] 
    \centering
    \includegraphics[width=1\linewidth]{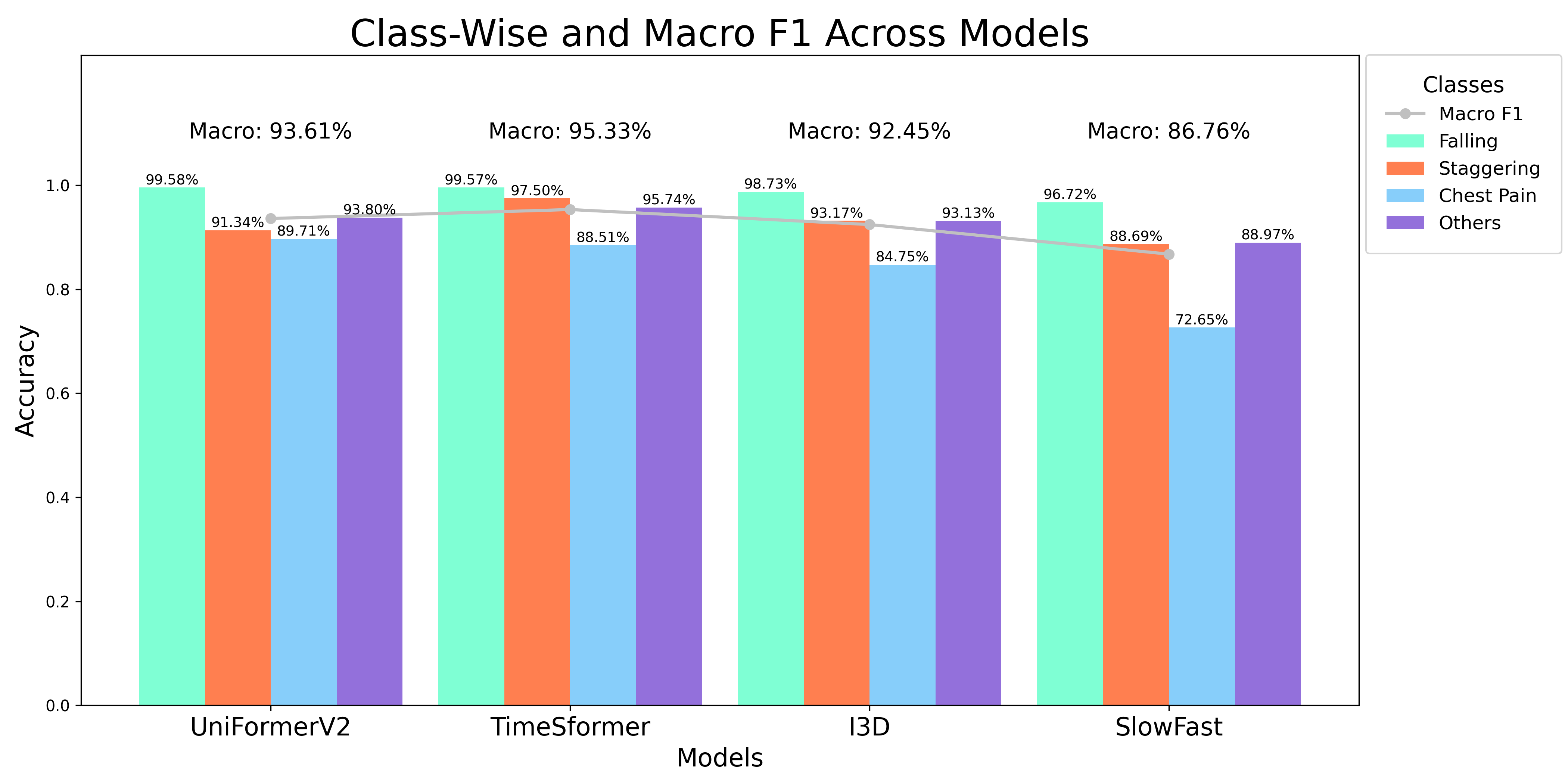}
    \caption{F1 Scores on Test Data}
    \label{fig:F1}
\end{figure}

A comparison of the four models by class-wise and macro F1 score is plotted in Figure \ref{fig:F1} for clearer visualization. Macro F1 score is the average of F1 scores from all classes. These values are between 0 and 1, with higher values representing better performance.

The graph clearly demonstrates that, class-wise, the F1 scores for predicting falls were promising using all these models, with the top three being UniFormerV2 (99.58\%), TimeSformer (99.57\%), and I3D (98.73\%). For predicting staggering, TimeSformer showed the best performance (97.50\%), followed by I3D (93.17\%), and UniFormerV2 (91.34\%) and SlowFast (88.69\%). When predicting chest pain, UniFormerV2 achieved highest F1 Score (89.71\%), followed by TimeSformer (88.51\%), I3D (84.75\%), and SlowFast (72.65\%). 

In terms of Macro F1 score, TimeSformer (95.33\%) achieved the highest value, followed by UniFormerV2 (93.61\%) and I3D (92.45\%) with a slight decrease, and SlowFast, with a suboptimal performance (86.76\%). Overall, all models demonstrated strong performance in predicting falls, moderately strong performance in predicting staggering and relatively lower performance in predicting chest pain.

Finally, normal actions misclassified as falls, staggering, and chest pain and their corresponding number of misclassified instances, are presented in Table \ref{tab:misclassify} in the Appendix, in order to further investigate model behaviour and error patterns. This provides insights into the types of actions that models struggled to distinguish from these three critical classes and paves the way to improve model performance, especially for the staggering and chest pain classes, which showed weaker results, in future work. We could increase samples in those misclassified actions to allow models to learn more of these cases in future training. The results reveal that the models tend to misclassify dynamic movements such as "throwing", "kicking something", "jumping", and "hopping" as "staggering". The actions most frequently misclassified as "chest pain" were "nod head/bow", "reach into pocket", "sit down", and "put on shoe".

\subsection{Performance and Efficiency Trade-offs Across Models}

\begin{figure}[h]
    \centering
    \includegraphics[width=1\linewidth]{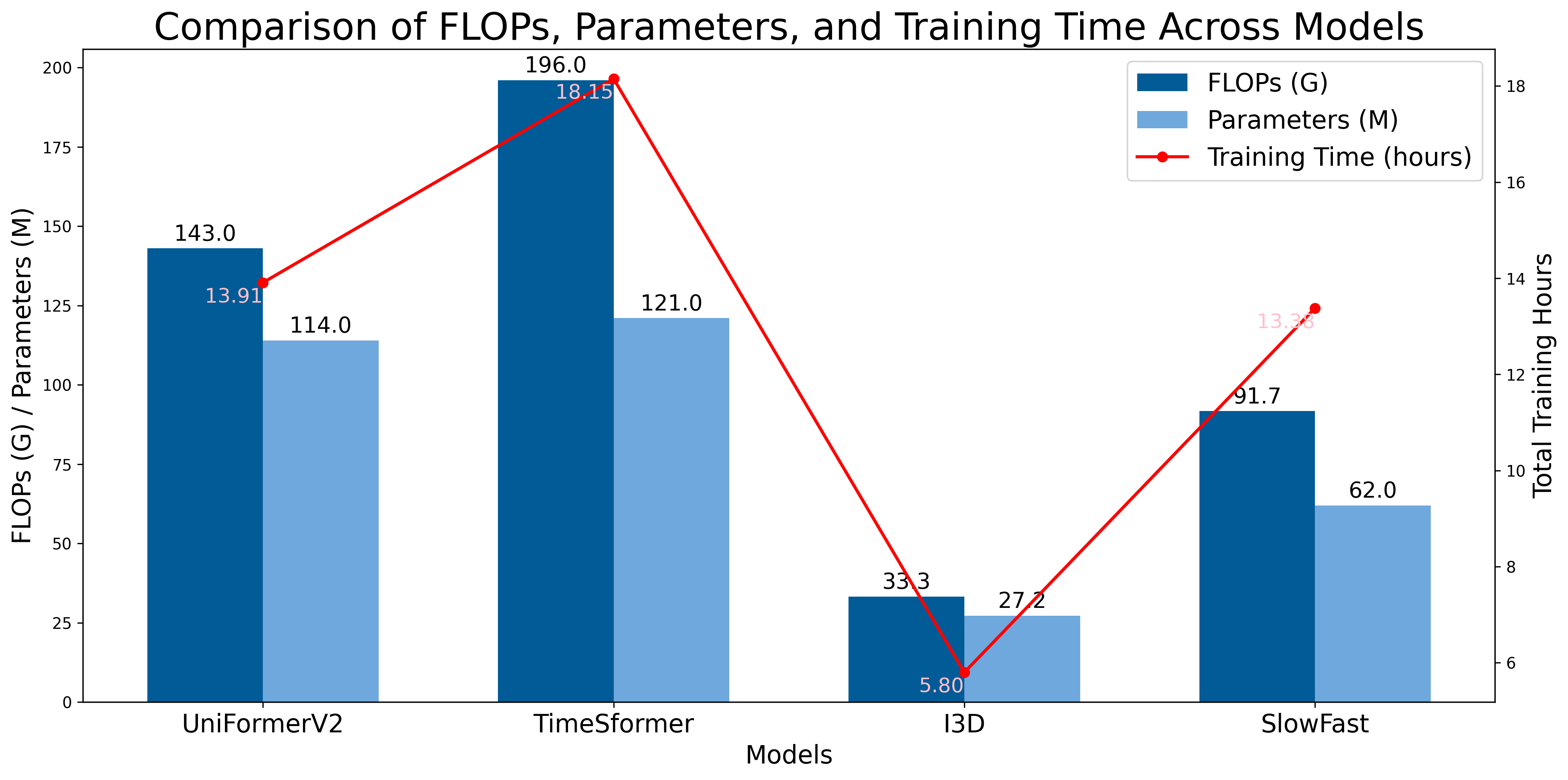}
    \caption{Comparison of FLOPs, Parameters, and Training Time}
    \label{fig:5}
\end{figure}

Figure \ref{fig:5} presents the comparison of FLOPs, number of parameters, and the training time. A positive correlation is seen between FLOPs, parameters and training time of these four models. Larger FLOPs and parameters require a longer training time. It is notable that the two models with attention components (UniFormerV2 and TimeSformer) are highly computationally intensive, with significantly more parameters and much longer training time compared to the other two. On the contrary, I3D has the lowest FLOPs, parameters and training time.

\begin{figure}[h]
    \centering
    \includegraphics[width=1\linewidth]{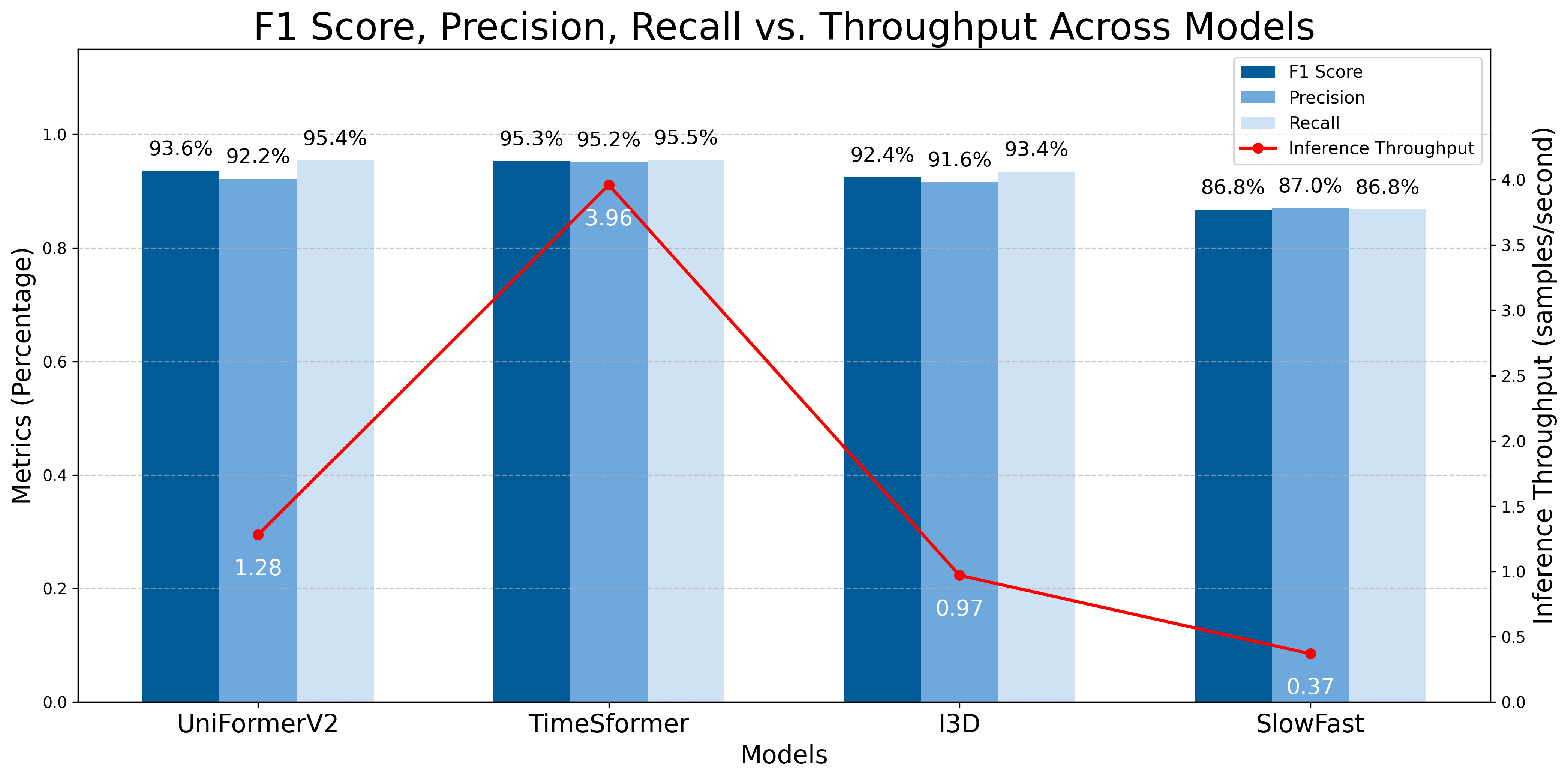}
    \caption{Comparison of Macro F1 Score, Precision, Recall, Against Throughput Across Models}
    \label{fig:6}
\end{figure}

Figure \ref{fig:6} shows the comparison of macro F1 score, macro precision, and macro recall vs. inference throughput to assess the relationship between model performance and inference efficiency. A model with higher values for all these metrics is preferred. TimeSformer demonstrated the fastest inference speed, tripling the performance of the model with the second rank, and had the top macro F1, recall and precision. UniFormerV2 sits in the second rank for macro recall, F1 score, and throughput. I3D ranked the third for these metrics. SlowFast showed the worst overall performance metrics and  slowest inference speed.  

\begin{figure}[h]
    \centering
    \includegraphics[width=1\linewidth]{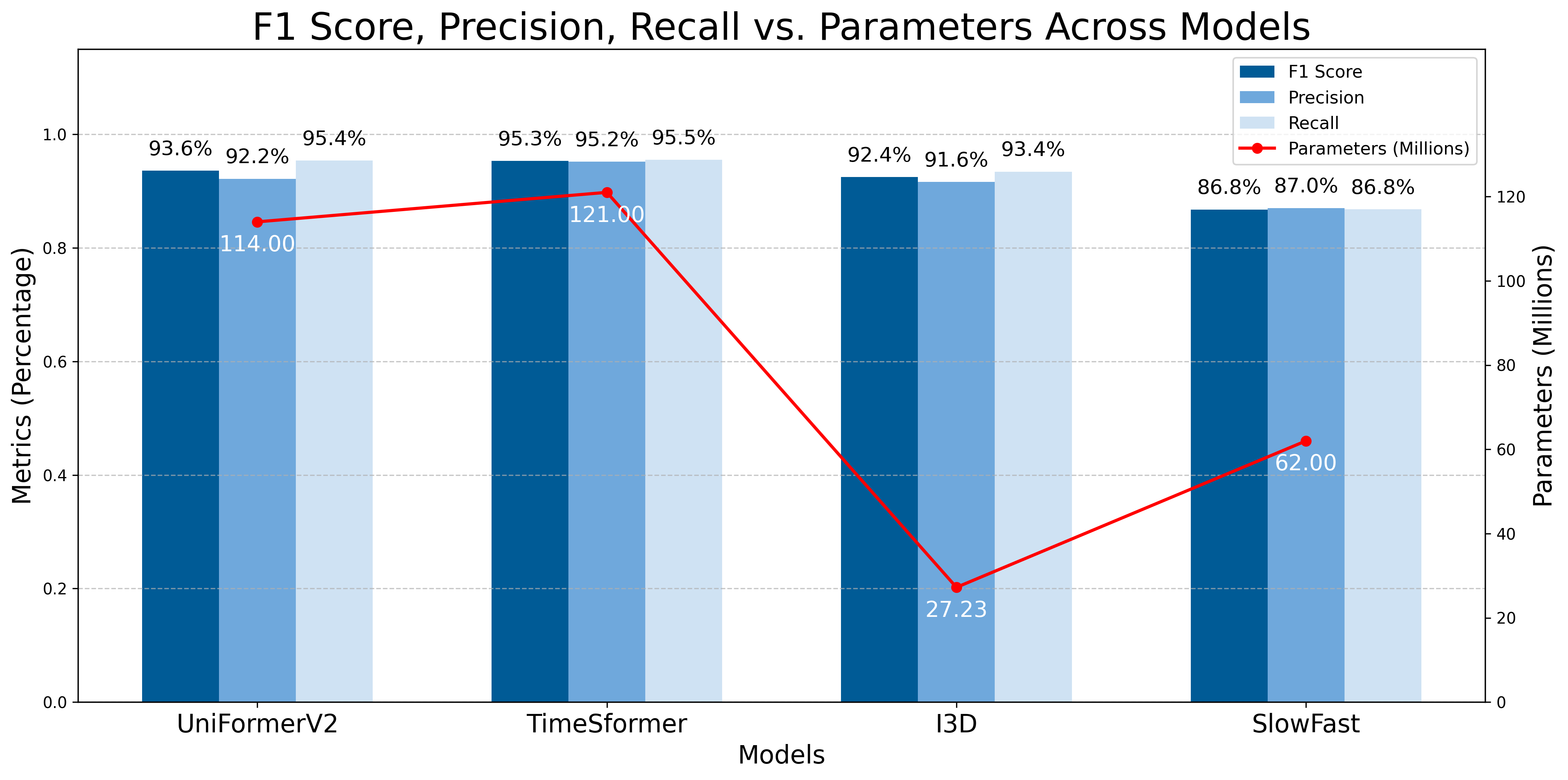}
    \caption{Comparison of Macro F1 Score, Precision, Recall, Against Parameters Across Models}
    \label{fig:7}
\end{figure}

Figure \ref{fig:7} compares the macro F1 score, macro precision, and macro recall against the number parameters to examine the relationship between the model performance and model complexity. It is not surprising that the two transformer-based models showed high performance with a larger number of parameters. But it is notable that I3D achieved strong performance with the smallest number of parameters.

\subsection{Discussion }

This study proposes a model to predict dangerous scenarios in assisted living. For real-life implementation, a model that exhibits high performance, high inference efficiency, low model complexity, and low computational cost is preferred. 

In conclusion, although some models may provide certain advantages in predicting specific classes, TimeSformer outperformed all models across all macro metrics for the three classes, achieving a macro F1 of 95.3\%, recall of 95.5\%, and precision of 95.2\%. In a real-time monitoring system, videos from all clients are queued for predicting; a faster inference throughput allows more samples to be processed within a limited time. Although TimeSformer requires longer total training time, its convergence speed is fast, and model training is much less frequent compared to inference, which may not need to be retrained for months. Conversely, inference runs 24/7 for assisted living. The high inference throughput of TimeSformer allows it to make inference for almost four times as many clients compared to I3D at the same time. Given that GPU servers on Amazon Cloud Service are billed hourly, high throughput significantly reduces the cost of the AI prediction system in the real-time HAR model in the long term. Moreover, our models were designed and implemented using GPU servers, therefore there are no hardware restrictions for training complex and computationally intensive models, such as TimeSformer. Ultimately, model selection is prioritised based on overall performance and inference throughput in our case, with TimeSformer proving to be the optimal model.

In contrast, the I3D model has the fewest parameters, FLOPs and training time, but still demonstrated good performance, slightly less promising than TimeSformer and UniFormer, with a macro F1 score of 92.5\%, recall of 93.4\%, and precision of 91.6\%. Despite slower inference speed compared to TimeSformer and UniFormer, I3D’s strong accuracy and low resource usage could potentially make it a strong candidate for resource-constrained environments.

\section{Conclusion and Future Work}

The aim of this research was to investigate to what extent a human action recognition (HAR) model can predict life-threatening scenarios such as falls, staggering and chest pain to improve assisted living environments. This research proposes a real-time human action recognition model that combines a video-based deep learning HAR model and a live video prediction and alert system. The results prove that transfer learning with state-of-the-art HAR models possesses a strong capacity to predict critical scenarios from daily actions. Notably, TimeSformer demonstrated a 99.57\% F1 score in predicting falls and a 97.50\% F1 score in predicting staggering. UniFormerV2 showed a 99.58\% F1 score in predicting falls and an 89.71\% F1 score in predicting chest pain. Overall, TimeSformer is a compelling candidate, as it achieved the highest macro F1 score (95.33\%) among all tested models, reflecting strong and balanced overall performance across different classes. It also achieved the highest macro recall (95.49\%), minimizing the risks of missing a critical event, and the best macro precision (95.19\%), presenting a low likelihood of producing false alarms. Moreover, TimeSformer demonstrated the fastest inference speed. Although it has higher model complexity and computational cost, these are not restrictions for deployment using GPUs, as suggested by this research. 

One limitation of this study is that the NTU RGB+D 60 dataset utilized is not specifically designed for assisted living environments, therefore its performance may require further testing in real-world settings; the dataset contains 40 daily action classes, which may not encompass all daily activities occurring inside assisted living, potentially limiting the model’s robustness when encountering unseen actions. Moreover, the paper has not benchmarked itself against sensor-based or skeleton-based HAR models for predicting these scenarios. Another limitation is that TimeSformer has high computational requirements which may make model training and deployment difficult with a much larger dataset size for real-life application development.  

AI-enabled real-time prediction of critical scenarios assists in immediate danger detection, potentially enhancing safety, trust, and comfort in assisted living. It allows the residents to maintain their independence with greater confidence, knowing that they are well protected, and that rapid intervention is possible during emergencies. This work could contribute to the Sustainable Development Goals (SDGs) by supporting health and well-being, sustainable communities and cities, and industrial innovation and resilient infrastructure. It might be improved by testing more deep learning HAR models. It is also worth experimenting with a skeleton-based HAR approach, which may be efficient in capturing human actions, and less resource-intensive. Moreover, it is suggested to include more critical scenarios, such as choking and vomiting, to create a more comprehensive model for predicting a series of dangers in assisted living. Furthermore, it is recommended to evaluate the model’s performance, if possible, using live video from real assisted living environments.

\bibliographystyle{ACM-Reference-Format}
\input{main.bbl}

\appendix

\section{Breakdown of Misclassified Normal Actions and Their Frequency}

Normal actions misclassified as falls, staggering and chest pain and their corresponding number of instances for the four models, are presented in Table \ref{tab:misclassify}. The action list is ordered in descending order of instance occurrence, followed by alphabetical order of words. 

\begin{table*}
  \caption{Normal Actions Misclassified as Falls, Staggering, and Chest Pain in Test Data}
  \label{tab:misclassify}
  \centering
  \renewcommand{\arraystretch}{1.3} 
  {%
  \begin{tabular}{lllll}
    \toprule
    Class & UniFormerV2 & TimeSformer & I3D & SlowFast \\
    \midrule
    Falls & throw 1 &  & throw 1 & pick up 2 \\
            &         &  &         & put on a shoe 1 \\   
    \midrule
    Staggering & jump up 5 & hopping 2 & hopping 3 & throw 1 \\
               & hopping 4 & throw 1   & throw 3   & sit down 1 \\
               &kicking something 4 &	put on a shoe 1 &	jump up 2	&stand up 1\\
               &put on a shoe 2&		&kicking something 2	&take off jacket 1 \\
               &throw 2	&	& drop 1 &	kicking something 1 \\
               &pick up 1& &		phone call 1 & \\	 
               & sit down 1	& &	pick up 1	& \\
               & stand up 1	& &	sit down 1  & \\	
                	& & &	stand up 1	& \\
    \midrule
    Chest Pain & nod head/bow 6&	nod head/bow 4&	reach into pocket 4	&reach into pocket 5 \\
   & reach into pocket 4	&put on glasses 2	&sit down 2	&put on a shoe 5\\
  &  put on a shoe 1	&reach into pocket 2	&drink water 1	&sit down 4 \\
  &  put palms together 1&	play with phone/tablet 1&	kicking something 1	&drink water 3 \\
   & salute 1&	put on a shoe 1&	nod head/bow 1&	point to something 3 \\
   & shake head 1	&sit down 1	&phone call 1&	nod head/bow 2 \\
   & sit down 1&	take off a shoe 1	&point to something 1 &	taking a selfie 2 \\
   & wipe face 1&	throw 1&	put on a hat/cap 1&	take off a shoe 2 \\

& &   & put on a shoe 1	      &  reading 2 \\
& &    &put on glasses 1	  &  drop 1  \\
& &   & put palms together 1 &	eat meal 1  \\
& &    &take off a shoe 1	 &   phone call 1  \\
& &   & take off jacket 1	  &  play with phone/tablet 1  \\
& &   & type on a keyboard 1	&put on glasses 1  \\
& &   &                  	&shake head 1  \\
& &   &                      	&type on a keyboard 1  \\
& &   &                      	&wipe face 1  \\
& &   &                      	&writing 1  \\
        \bottomrule
  \end{tabular}%
  }
\end{table*}

\end{document}